\documentclass{article}

\usepackage{PRIMEarxiv}

\usepackage[utf8]{inputenc} 
\usepackage[T1]{fontenc}    
\usepackage{hyperref}       
\usepackage{url}            
\usepackage{booktabs}       
\usepackage{amsfonts}       
\usepackage{nicefrac}       
\usepackage{microtype}      
\usepackage{lipsum}
\usepackage{fancyhdr}       
\usepackage{graphicx}       
\graphicspath{{media/}}     

\title{Enhancing autonomous vehicle safety in rain: a data-centric approach for clear vision
}

\author{
  Mark A. Seferian and Jidong J. Yang* \\
  Smart Mobility and Infrastructure Laboratory \\
  College of Engineering\\
  University of Georgia \\
  Athens, GA, 30602, USA\\
  \texttt{\{Mark.Seferian@uga.edu, Jidong.Yang@uga.edu\}} \\
}

\begin{document}
\maketitle

\begin{abstract}
Autonomous vehicles (AV) face significant challenges in navigating adverse weather, particularly rain, due to the visual impairment of camera-based systems. In this study, we leveraged contemporary deep learning techniques to mitigate these challenges, aiming to develop a vision model that processes live vehicle camera feeds to eliminate rain-induced visual hindrances, yielding visuals closely resembling clear, rain-free scenes. Using the Car Learning to Act (CARLA) simulation environment, we generated a comprehensive dataset of clear and rainy images for model training and testing. In our model, we employed a classic encoder-decoder architecture with skip connections and concatenation operations. It was trained using novel batching schemes designed to effectively distinguish high-frequency rain patterns from low-frequency scene features across successive image frames. To evaluate the model’s performance, we integrated it with a steering module that processes front-view images as input. The results demonstrated notable improvements in steering accuracy, underscoring the model’s potential to enhance navigation safety and reliability in rainy weather conditions.
\end{abstract}


\section{Introduction}
Rain poses significant challenges to not only human visual perception, but also for autonomous vehicles (AV) navigating roadways. Rain streaks can severely hinder camera-based objects and feature detection systems employed by AV. Consequently, automotive manufacturers often deactivate autonomous driving features during inclement weather. In response, research within the field of developing deraining deep learning models has seen a surge of interest in recent years. However, due to the intricate nature of dynamic rain streaks and slowly changing background scenes, deraining remains a challenging task.

We aim to address the deraining challenge for AV by: (1) Developing a deep learning based vision model capable of removing rain streaks, yielding results that resemble a clear, rain-free image, (2) using a data centric approach to devise and analyze different batching schemes to enhance model training and inference performance, and (3) utilizing an established steering angle predication model to validate the benefits of deraining in improving AV’s steering performance.

The datasets for this study are generated from the Car Learning to Act (CARLA) simulator. Three datasets are prepared for model training, validation, and testing purposes. Each dataset comprises of rainy images as the input and corresponding clear images as the label. The training dataset includes diverse maps and environments to improve the model’s generalization. Moreover, the validation and testing datasets are derived from maps not used in the training dataset. Deraining, akin to denoising, is a common machine learning task, exemplified by methods like denoising autoencoder \cite{vincent2008extracting}.

However, the challenge of removing rain streaks from an ego vehicle's camera view differs significantly from denoising static images due to the dynamic nature of sequential scenes captured by these cameras. Here, rain streaks act as high-frequency signals (similar to noises), contrasting with the low-frequency signals of camera scenes, which evolve slowly as the vehicle navigates roads. Removing rain streaks while maintaining the integrity of scenes is analogous to filtering out high-frequency signals. In order to harness the distinct dynamics of these signal types, two novel batching schemes are employed and compared to the conventional batching to assess their impact on the model training and the resultant deraining performance. The first batching scheme uses paired images that are sequential in time and sequential in batch (STSB), while the second batching scheme uses paired images that are sequential in time and random in batch (STRB). In contrast, the traditional batching scheme relies on image pairs that are random in time and random in batch (RTRB), where the temporal cue is lost within each batch.

The model architecture devised in this study draws inspiration from two influential architectures: The Deep Convolutional Generative Adversarial Network (DCGAN) \cite{radford2016unsupervised} and the U-Net \cite{ronneberger2015unet}. DCGAN, renowned for its applications in computer vision such as image generation \cite{karras2017progressive,brock2019large,zhang2016stackgan}, style transfer \cite{karras2021stylebased,xu2021drbgan}, and data augmentation \cite{zhen2023cosupervised,motamed2021data,jadli2020dcgan}, serves as a foundational pillar in our approach. Moreover, U-Net, initially developed for biomedical image segmentation, is important in modern diffusion models for iterative image denoising \cite{ramesh2021zeroshot,rombach2022high,xia2023diffir}. To tackle the challenge of image deraining, we propose an encoder-decoder architecture that extends the DCGAN to accommodate higher image resolutions while integrating the skip-concatenation mechanism from U-Net to leverage multiscale perceptual views, fostering context-aware denoising.

To illustrate the effectiveness of our model, we compare the derained images against both rainy and ground-truth (clear) images. Additionally, we benchmark our model against PreNet \cite{ren2019progressive}, a seminal work in the field. To quantitatively assess the performance of our deraining model, we employ PilotNet \cite{bojarski2016end}, a steering angle predictor. Steering performance is evaluated under rainy, clear, and derained conditions to demonstrate the advantages of our model in improving vehicle steering under adverse weather conditions. In summary, the key contributions of this study are as follows:
\begin{itemize}

\item We introduce sequential batching schemes that facilitate cost-free learning of structured scenic features against noisy rain streaks. This data-centric approach enhances both training stability and inference performance.

\item Inspired by DCGAN and U-Net, our proposed simple yet effective architecture surpasses the prior work in removing rain streaks from images.

\item The efficacy of our deraining model is validated through steering performance evaluation using PilotNet, where steering angles predicted from derained images closely match those from clear images.
\end{itemize}

\section{Deraining}
A number of deep learning models have been proposed to address the deraining challenge. DID-MDN \cite{zhang2018density} utilized a multi-dilation network to capture rain streaks of varying sizes using dilated convolutions to capture long-range dependences in rain streak patterns. JORDER \cite{yang2016deep} jointly addresses rain detection and removal within a unified framework by extracting rain discriminative features. PReNet \cite{ren2019progressive} used a modified progressive residual network (PRN) to remove rain streaks by progressively refining the deraining results through multiple stages. DerainNet \cite{fu2017clearing} employed a deep convolutional neural network (CNN) \cite{lecun1989backpropagation} to directly learn the mapping relationship between rainy and clear images. Restormer \cite{zamir2022restormer} used a multiscale hierarchal design incorporating efficient transformer blocks, such as multi-Dconv head transposed attention and gated-Dconv feed-forward network to derain images. KBnet \cite{zhang2023kbnet} argued against transformer models as they lack desirable inductive bias of convolutions. Instead, it incorporated a kernel basis attention model to adaptively aggregate spatial information and a multi-axis feature fusion to encode and fuse diverse features for image restoration.

Other researchers in the field have adopted GAN \cite{goodfellow2020generative} based architectures for image deraining. DerainCycleGAN \cite{wei2021deraincyclegan} used an unsupervised attention guided rain streak extractor, two generators, and two discriminators to derain images. ID-CGAN \cite{zhang2019imageraining} integrated skip-connections and DenseNet Block that uses per-pixel loss and perceptual loss to improve deraining performance. PAN \cite{wang2018perceptual} employed a perceptual adversarial loss and hidden trainable layers. FS-GAN \cite{xiang2019single} incorporated feature-supervision on generator layers to contribute gradient information for optimization to improve image deraining. IGAN \cite{ren2020single} followed a divide-and-conquer strategy to divide image deraining into rain locating, removal, and detail refinement sub-tasks.

In contrast to researchers who focused on model architecture, we emphasize a data centric approach using cost-free batching schemes to improve image deraining performance. For proof of concept, we devise a simple encoder-decoder architecture with end-to-end training for direct image deraining and style transfer.

\section{Data collection}
Data collection and curation is pivotal in our data-centric approach, profoundly shaping the training of our model. We aim to achieve three objectives: (1) Capturing both rainy images and their corresponding clear counterparts as an ego vehicle navigates roads, facilitating direct end-to-end training with sequential images; (2) ensuring diversity in the driving environments captured within the datasets; and (3) acquiring steering wheel angle data alongside image data to enable quantitative evaluation of deraining on steering performance.

\subsection{CARLA simulator}
\begin{figure}
    \centering
    \includegraphics[width=0.8\linewidth]{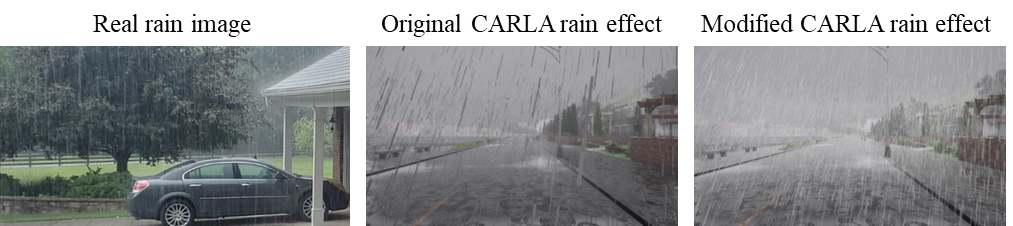}
    \caption{Visualization of heavy rain effect when compared to original CARLA rain effect and modified CARLA rain effect.}
    \label{fig:Fig1}
\end{figure}

To collect necessary data for model training and testing, Car Learning to Act (CARLA) \cite{dosovitskiy2017carla}, a simulator for autonomous driving research, is utilized. CARLA is a powerful open-source simulator that contains various digital assets, such as vehicles, sensors to capture data, and pre-made maps that include a diverse selection of environments. CARLA also has an extensive API, offering flexibility in setting the time of day, controlling weather conditions, and gathering necessary vehicle data. However, one glaring issue with CARLA is its simulated rain effects. The original rain effects in CARLA based on the 0.9.14 release were unrealistic when compared to real rain. Consequently, modifying the rain effects in CARLA to closely reflect real-world rainy conditions becomes necessary. To modify the rain effects, a custom-built version of CARLA is created using the Unreal Editor to modify the rain asset file to reflect real-world rain effects. Figure \ref{fig:Fig1} shows a comparison of a real-world heavy rain image, the original CARLA heavy rain image, and the modified CARLA heavy rain image.

The rain streaks in the real-world image can be described as thin with a light grey color, while the original CARLA rain effect is drastic in comparison. The rain streaks are thick rectangular pixels, with a dark grey color to them. After modifying the rain asset file, the rain streaks are thinner, lighter in color, closely resembling real-world rain effect.

\subsection{Image data}
The datasets for the model training required a rainy input image and a corresponding clear label image. It was essential that the simulation runs were synchronized to capture the clear and rainy images at the exact frame. To synchronize the frames, a Python script was developed to ensure that the ego vehicle followed a predefined path in CARLA. This script sets the simulation in synchronous mode to manually call the simulation to move forward a time step. This ensures that frames across different simulation runs were identical. The forward-view images were captured through the CARLA spectator view that was attached to the hood of the ego vehicle. In the simulation runs, no other moving vehicles were present, and all traffic lights were set to green when the ego vehicle approached to prevent repetitive, standstill images for extended periods of time. The time of day was set to Noon for generating both the clear and rainy road scene images.

\begin{figure}
    \centering
    \includegraphics[width=0.7\linewidth]{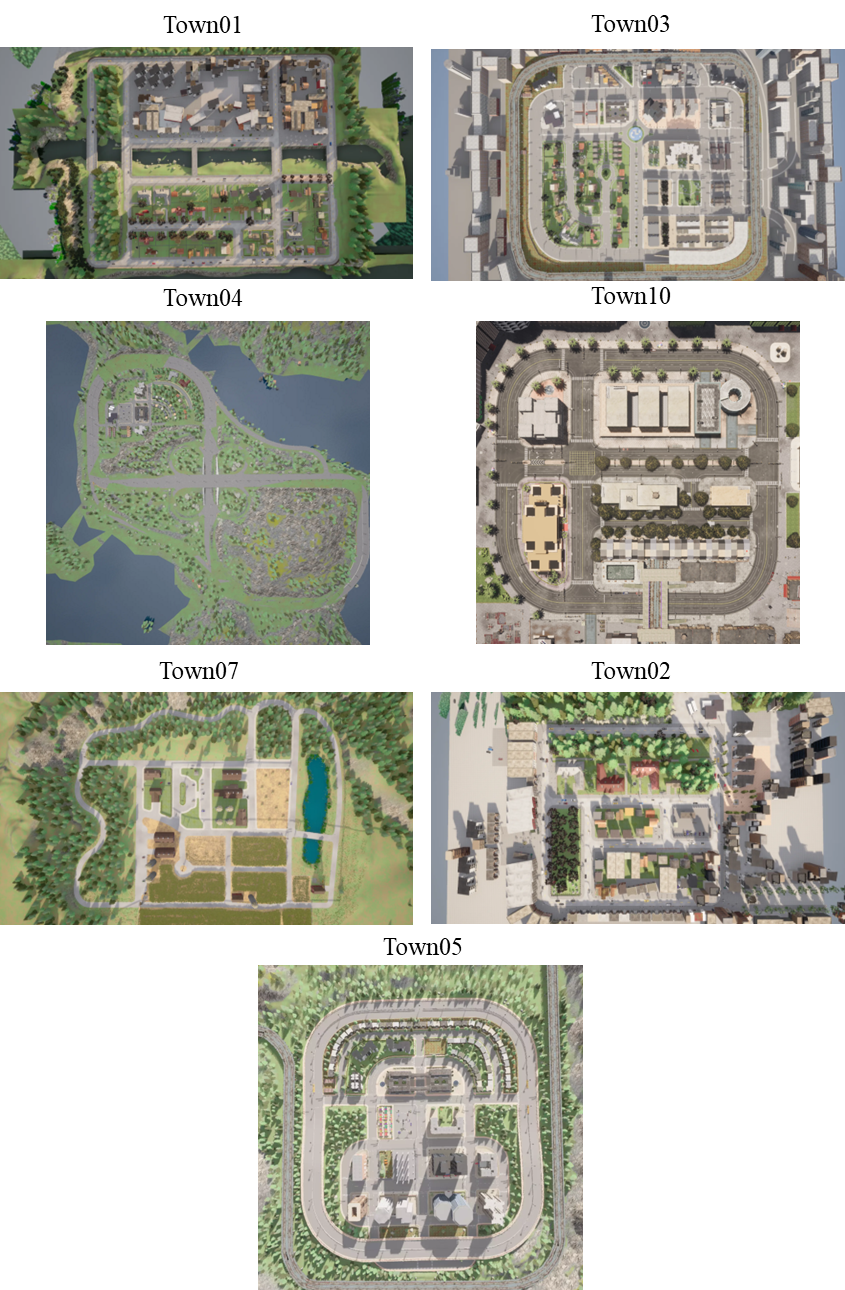}
    \caption{Aerial views of the CARLA maps used. Town01, 03, 04, 07, and 10 for training; Town02 for validation; and Town05 for testing.}
    \label{fig:Fig2}
\end{figure}

CARLA offers various pre-made maps that can be used to create image datasets containing diverse road scenes. The aerial views of the maps used are shown in Figure \ref{fig:Fig2}.

For the training datasets, the following five distinct maps were included: Town01, Town03, Town04, Town07, and Town10. Town01 featured a small river surrounded by a mix of commercial and residential buildings in a forest terrain. Town03 was an urban landscape with metro tracks, a blend of commercial and residential buildings, and a roundabout. Town04 offered highway roads winding through a mountainous terrain with an exit to a small town. 

Town07 presented a rural countryside setting with narrow winding roads, farming structures, and cornfields. Town10 provided a downtown environment with skyscrapers, residential complexes, and an ocean view. For the validation dataset, Town02 was selected. For the testing dataset, Town05 was chosen due to its larger size and comprehensive environmental characteristics, such as highways, residential and commercial zones, skyscrapers, tree-lined streets, and metro tracks. Consequently, the training dataset comprised 32,000 images (16,000 clear images and 16,000 rainy images). The validation dataset contained 3,200 images, while the testing dataset consisted of 4,000 images.

\subsection{Steering angle data}
To capture the steering angle data for each map in CARLA, we recorded the steering angle of the vehicle at each frame. It is worth noting that the CARLA API allowed only for capturing the drive wheel angles. However, the PilotNet model outputted steering wheel angles. Thus, a conversion from drive wheel angles (P) to steering wheel angles (S) was applied by multiplying the drive wheel angles by the steering wheel ratio (R). The resultant steering wheel angles for each map were saved along with the corresponding frame number. This data was used for assessing the effect of deraining on steering performance.

\section{Data-centric approach}
This study emphasizes the data-centric aspect rather than the model architecture design, highlighting the critical role of data batching in learning distinct signals at different frequencies to enhance autonomous vehicle vision.

\subsection{Data preparation and batching schemes}
Data preparation is vital, as it significantly influences model training. Initially, all images were center cropped and resized to 256x256. Normalization was omitted due to its adverse effect on the quality of derained images. The training process involved correctly pairing rainy input images with their corresponding clear images as ground truth. This was accomplished by creating two separate folders and using the frame number to correctly pair them together.

\begin{figure}
    \centering
    \includegraphics[width=0.8\linewidth]{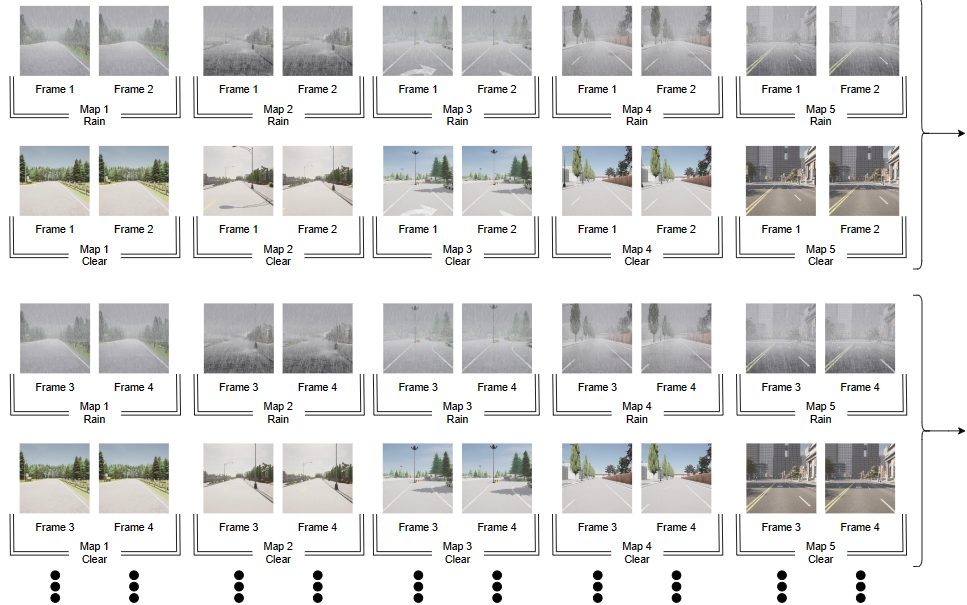}
    \caption{Batching Scheme 1: Sequential in time and sequential in batch (STSB).}
    \label{fig:Fig3}
\end{figure}

Three distinct batching schemes were implemented using a customized dataloader. The first scheme, termed Sequential in Time and Sequential in Batch (STSB), paired two sequential frames of rainy and clear images from each of the five training maps within a batch. This scheme operated sequentially both in time and batch, as illustrated in Figure \ref{fig:Fig3}, where Frames 1 and 2 of rainy and clear images from each map formed the first batch, followed by Frames 3 and 4 in the second batch. This process was repeated until all images were utilized. However, since not all maps contained an equal number of images, once the images from one map were depleted, another remaining map was randomly chosen to fill the batch. 

\begin{figure}
    \centering
    \includegraphics[width=0.8\linewidth]{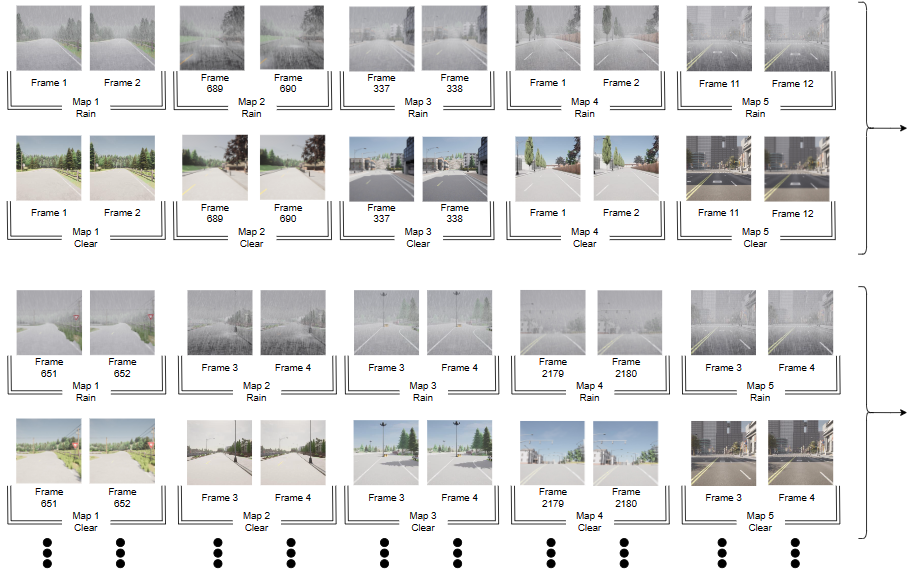}
    \caption{Batching Scheme 2: Sequential in time and random in batch (STRB).}
    \label{fig:Fig4}
\end{figure}

The second novel batching scheme, Sequential in Time and Random in Batch (STRB), maintained the paring of two sequential frames but introduced randomness in batch loading. This meant that within each batch, the frames were shuffled randomly rather than following a strict sequential order. This is demonstrated in Figure \ref{fig:Fig4}, where sequential frames from each map are paired, but their order within each batch is randomized.

\begin{figure}
    \centering
    \includegraphics[width=0.7\linewidth]{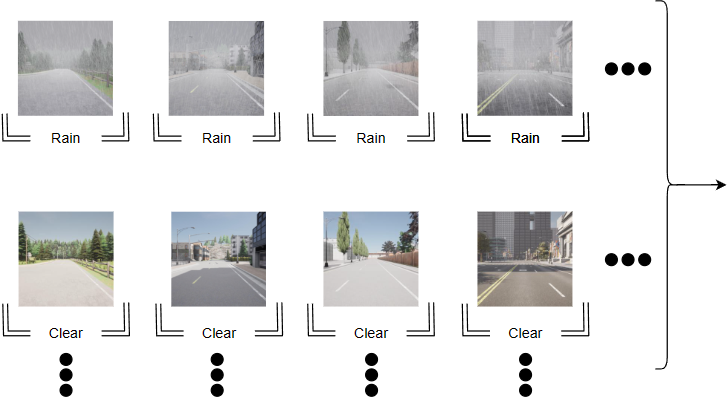}
    \caption{Batching Scheme 3: Random in time and random in batch (RTRB).}
    \label{fig:Fig5}
\end{figure}

The third batching scheme, Random in Time and Random in Batch (RTRB), represented conventional batching, as shown in Figure \ref{fig:Fig5}. In this scheme, both the frame pairs and their order within each batch were randomly selected.

\subsection{Model architecture} 
Our model architecture followed a classic encoder-decoder design, where the encoder and decoder were adapted from the discriminator and generator of DCGAN to enable higher resolution images. Particularly, the original DCGAN architecture handled 64x64 images. Additional convolution blocks were added to allow for 256x256 images. Skip-concatenation \cite{ronneberger2015unet} was adopted between convolutional blocks with same spatial resolution in the encoder and the decoder. This design enabled direct flow of hierarchical, multiscale features from the encoder to the decoder, enabling context-aware image generation, which was crucial in tasks like deraining, where understanding the fundamental structure of the road scene was essential for removing rain streaks and enhancing image clarity.

\begin{figure}
    \centering
    \includegraphics[width=0.8\linewidth]{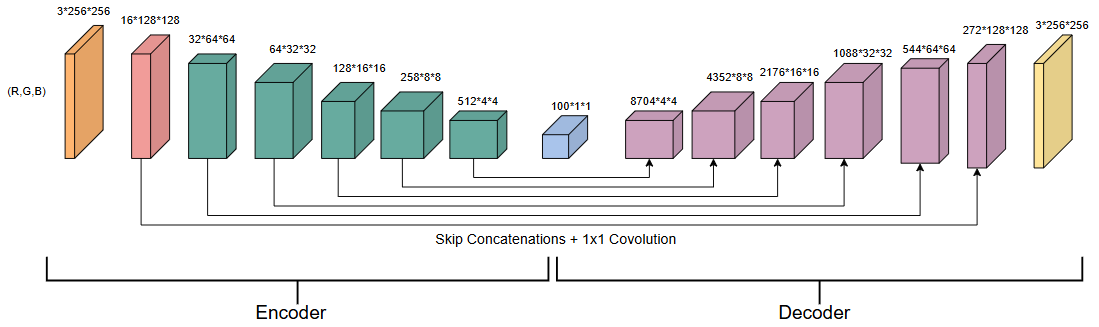}
    \caption{Model Architecture. The decoder transposed convolutions (in purple) are modified by concatenating with corresponding convolution block from the encoder, followed by 1x1 convolution to resize the channel dimension.}
    \label{fig:Fig6}
\end{figure}

\begin{figure}
    \centering
    \includegraphics[width=0.8\linewidth]{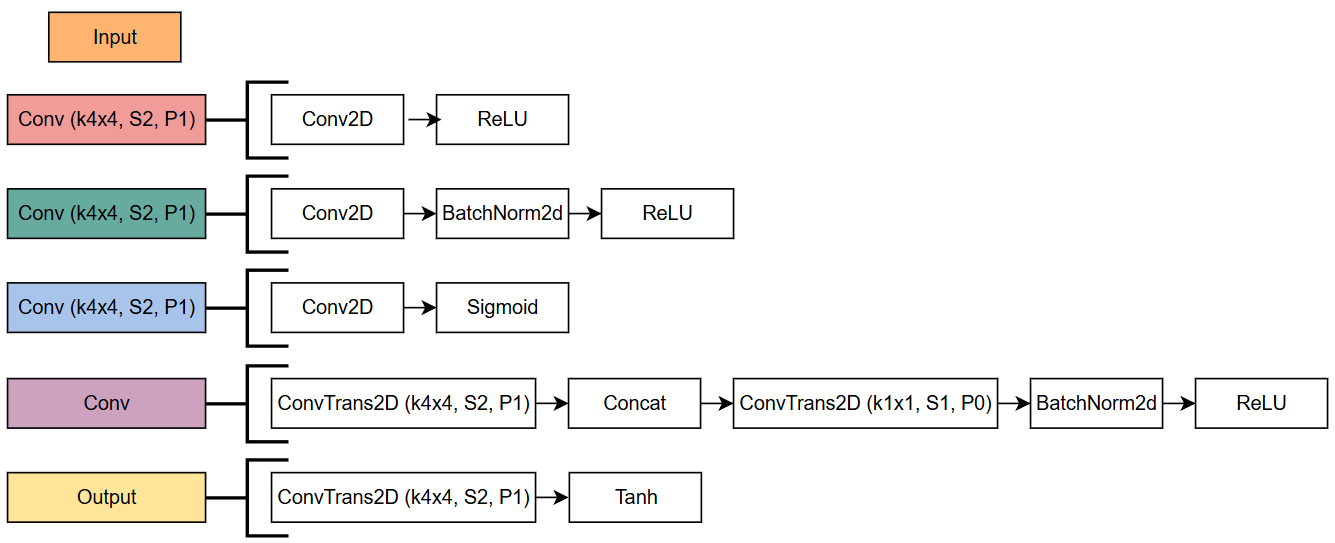}
    \caption{Computational modules of colored block in Figure \ref{fig:Fig6} (k: kernel size; S: Stride; P: Padding).}
    \label{fig:Fig7}
\end{figure}

Figure \ref{fig:Fig6} shows the proposed deraining model architecture with distinct blocks denoted by different colors. Figure \ref{fig:Fig7} further elaborates on computational details of each colored block. Batch Norm \cite{ioffe2015batch} was applied to all layers except the decoder output and encoder input. In line with the principle of design simplicity \cite{springenberg2014striving}, the model exclusively used convolutional layers, where down-sampling was achieved by increasing the stride. ReLU was predominately used as nonlinearity across convolution layers, while Tanh was used for the decoder output and sigmoid was employed for the encoder output.

\section{Model training and evaluation}

For each batching scheme, the deraining model was trained with 100 epochs, a batch size of 10, and a learning rate of 0.0002. We use MSE loss and Adam Optimizer \cite{kingma2014adam} with parameters $\beta_1$ = 0.5 and $\beta_2$ = 0.999. All experiments are conducted on a workstation using an AMD Ryzen 9 7950x CPU, 32GB of Ram, and Nvidia GeForce RTX 4090 24GB.

\subsection{Batching scheme performance}
\begin{figure}
    \centering
    \includegraphics[width=0.5\linewidth]{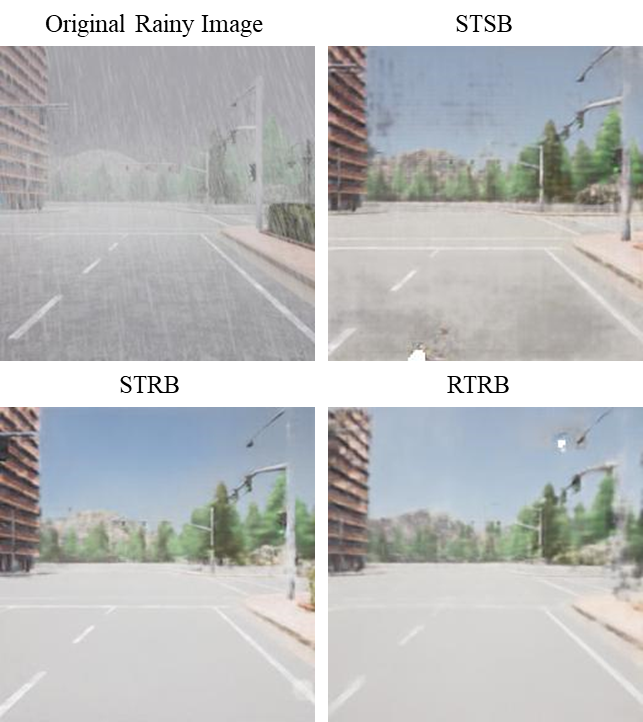}
    \caption{Visualization of deraining results of a single frame.}
    \label{fig:Fig8}
\end{figure}

Table \ref{tab1} summarizes train, validation, and test losses for different batching schemes.
As shown in Table \ref{tab1}, the STRB batching scheme performs the best, followed by RTRB and STSB. For visual comparison, Figure \ref{fig:Fig8} shows the derained images from the three batching schemes.
\begin{table}
\centering
\caption{MSE loss comparison of different batching schemes.}
\begin{tabular}{l l l l}
\hline
Batch Scheme & Train Loss & Validation Loss & Test Loss \\
\hline
STSB & 0.0122 & 0.0821 & 0.0132 \\

STRB & \textbf{0.0012} &\textbf{0.0130} & \textbf{0.0106} \\

RTRB & 0.0014 & 0.0164 & 0.0115 \\
\hline

Note: bold indicates the best performance.

\end{tabular}
\label{tab1}
\end{table}

\begin{figure}
    \centering
    \includegraphics[width=0.6\linewidth]{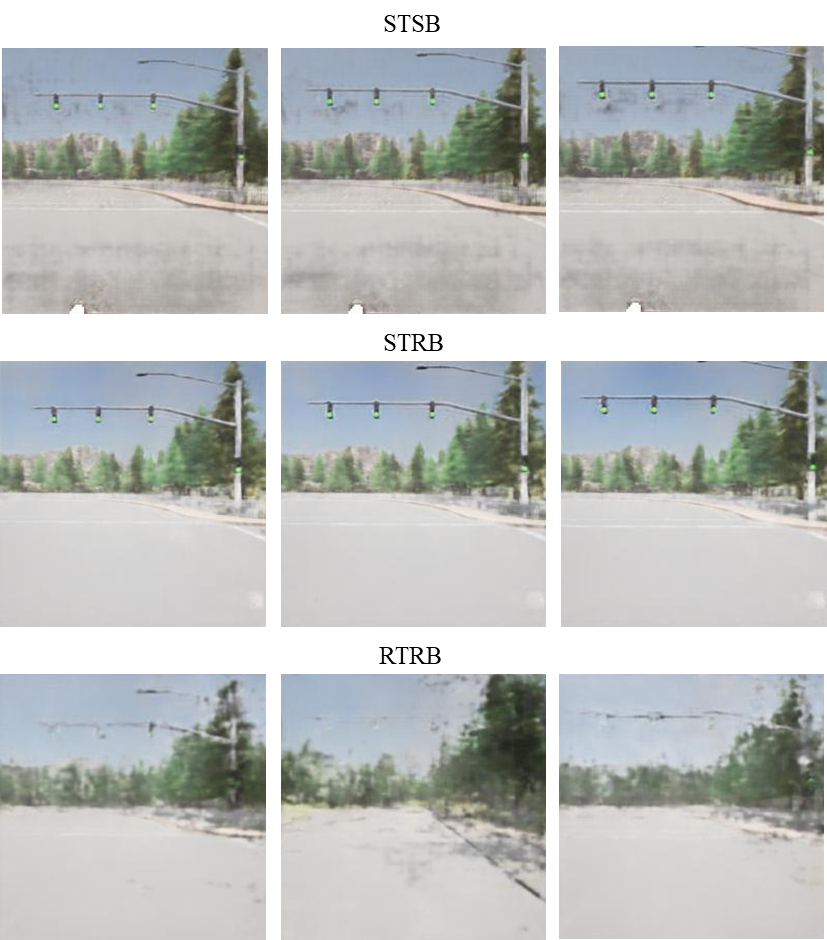}
    \caption{Visualization of deraining results of three consecutive frames.}
    \label{fig:Fig9}
\end{figure}

Notably, STSB has lingering grey and white spots in the sky and pavement areas, where solid color or gradient of color are expected. In contrast, RTRB shows improvement over STSB, with the absence of grey spots. However, some artifacts (e.g. a white spot) exist in the sky area and structural information (e.g., the light post) is lost. STRB, on the other hand, performs extremely well in comparison, preserving both pixel-level information as well as structural features. For further comparison, three consecutive derained images for each of the three batching schemes are shown in Figure \ref{fig:Fig9}.

It becomes apparent that RTRB struggles with slight environmental movements as the car navigates down the road. Also, the overall image quality exhibits watery visuals with significant loss of details, especially in object structures, such as traffic lights and trees. For STSB, various grey spots are present in the image, which likely arise from less diverse backgrounds due to sequential batching. In contrast, STRB harnesses the advantages of both sequential frames and random batching, resulting in improved images with pixel-level and structural integrity.

In summary, STRB is a novel batching scheme that utilizes random batching of sequential frames to derain images. This strategy enables the model to effectively capture the distinct dynamics of raindrops against slowly changing roadway scenes, resulting in superior deraining performance when compared to the traditional RTRB approach. By using sequential frames, STRB can better understand the rain dynamics between the consecutive frames to adaptively remove rain streaks while preserving the scene details and integrity. On the other hand, the randomness in the batch increases diversity in scenes within each batch, mitigating overfitting and bias toward any particular scenes. As such, the STRB batching scheme effectively preserves both structural and pixel-level details when deraining images. 

\subsection{Comparison of deraining results}
\begin{figure}
    \centering
    \includegraphics[width=0.5\linewidth]{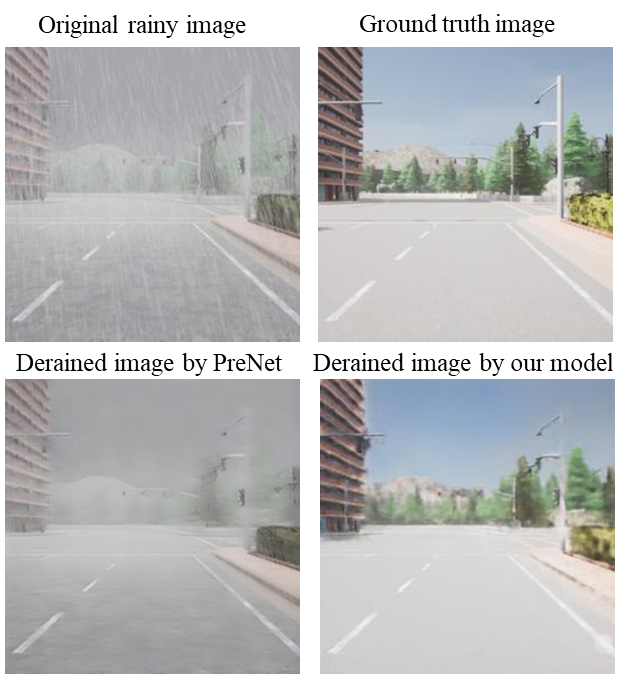}
    \caption{Scene 1: AV approaching traffic lights.}
    \label{fig:Fig10}
\end{figure}

\begin{figure}
    \centering
    \includegraphics[width=0.5\linewidth]{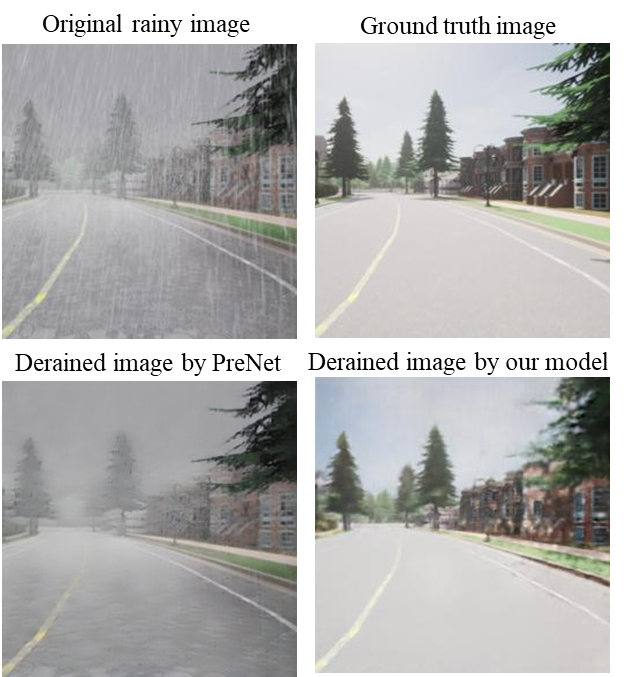}
    \caption{Scene 2: AV driving through residential area.}
    \label{fig:Fig11}
\end{figure}

\begin{figure}
    \centering
    \includegraphics[width=0.5\linewidth]{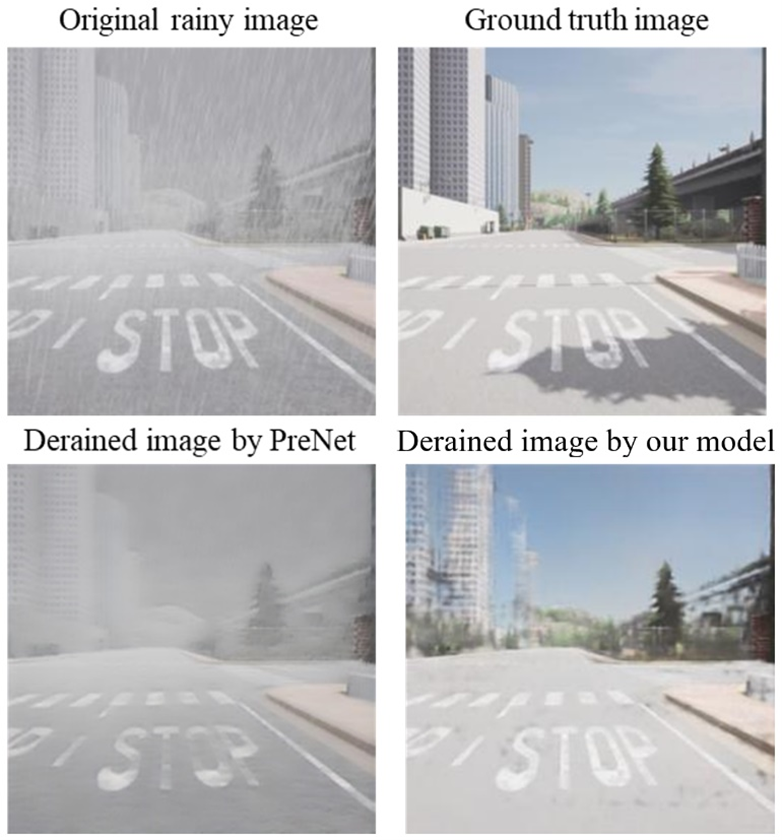}
    \caption{Scene 3: AV approaching 4-way stop.}
    \label{fig:Fig12}
\end{figure}

With the STRB batching scheme demonstrating the highest effectiveness for deraining, we compare our model’s deraining performance to PreNet, the baseline model used in this study. For a qualitative assessment, three distinct roadway scenes are used for comparison. For each scene, four images are presented, including the original rainy image, the ground-truth image, the PReNet derained image, and the derained image from our work. As shown in Figures \ref{fig:Fig10}, \ref{fig:Fig11}, and \ref{fig:Fig12}, our model with the STRB batching scheme consistently outperforms PReNet.

While PReNet successfully removes the rains streaks from the rainy image, the resulting image quality drastically decreases. Additionally, the environment retrains the original gray, colorless appearance of the rainy image. Our model, on the other hand, has not only removed the rain streaks of the image, but also retained the environmental details to mimic a clear sunny day. As a result, the derained image views from our model show much greater visibility than those from PReNet. It is important to note that our model was trained to learn direct mapping between rainy images and corresponding clear, sunny images. Implicitly, our model learns to tackle two tasks simultaneously: (1) Removing rain streaks and (2) style-transferring from rainy weather conditions to clear and sunny weather conditions.

When comparing the derained image to the ground-truth, there is a slight loss of detail, most notably in the leaves of trees in the second scene. However, the overall image quality remains sufficient for driving-relevant feature and object detections, such as the lane markings, the “STOP” text on the road, and roadside structures are visible. One area where the model struggles is with skyscrapers (as seen in Scene 3 of Figure \ref{fig:Fig12}), where some segments of the building are missing or distorted. Despite these, the model performs remarkably well at deraining images, representing a leap forward compared to the prior work. It is important to emphasize that our primary objective of this study is to mitigate the adverse effects of rain while ensuring that essential features remain visible for real-time driving tasks rather than achieving a perfect high-resolution reconstruction of all scene details. While the latter could potentially be addressed by scaling up the network with architectural enhancements, such improvements would increase computational costs and fall outside the scope of this study. 

\subsection{Steering performance}
To quantify the benefits of image deraining achieved by our model, PilotNet was employed to predict steering angles for clear, rainy, and derained images. Since PilotNet was originally designed for lane-following tasks, scenarios involving intersections and sharp turns were excluded from the analysis. The evaluation was conducted on a multi-lane highway comprising straight segments and gradual turns. As previously noted, the ground-truth steering angles were directly recorded from CARLA.

It is important to acknowledge that even for clear images, PilotNet exhibits an inherent deviation from the ground-truth steering angles recorded in CARLA. For steering performance evaluation, the mean absolute error (MAE) was computed to measure the deviation under four conditions: Clear weather, heavy rain, light rain, and derained images. Table \ref{tab2} presents the results, indicating an inherent error of 0.356 degrees for clear weather and a slightly higher error of 0.508 degrees under derained conditions. In comparison, both heavy and light rain conditions result in larger steering errors, with heavy rain showing a significantly worse performance. Notably, the steering error under heavy rain is more than twice that observed for light rain.

\begin{table}
\centering
\caption{Mean absolute steering angle error.}
\begin{tabular}{l l}
\hline
\textbf{Condition} & \textbf{Error (degree)} \\
\hline
Clear & 0.356 (Inherent Error) \\
Heavy Rain & 1.204 \\
Light Rain & 0.561 \\
Derained & 0.508 \\
\hline
\end{tabular}
\label{tab2}
\end{table}

\begin{figure}
    \centering    \includegraphics[width=0.7\linewidth]{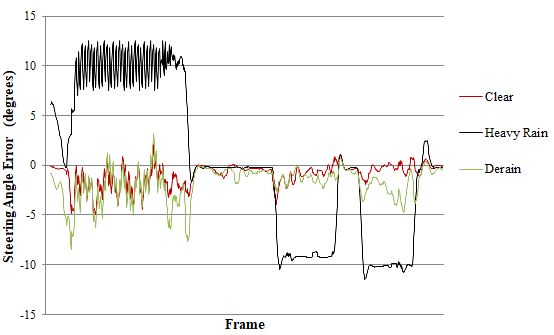}
    \caption{Steering angle comparison error (heavy rain).}
    \label{fig:Fig13}
\end{figure}

\begin{figure}
    \centering    \includegraphics[width=0.7\linewidth]{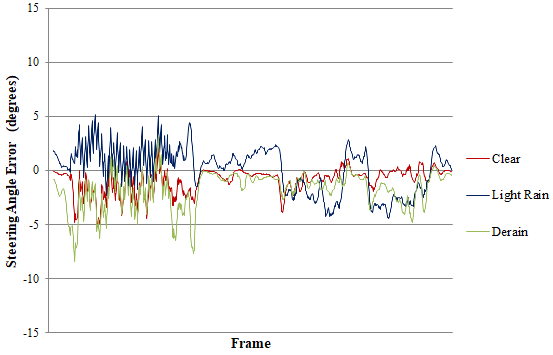}
    \caption{Steering angle comparison error (light rain).}
    \label{fig:Fig14}
\end{figure}

Figures \ref{fig:Fig13} and \ref{fig:Fig14} illustrate the live steering angle error relative to the ground truth over a simulation run for three scenarios: Clear, heavy rain, and derained. As shown in Figure \ref{fig:Fig13}, the heavy rain scenario exhibits three segments of significant deviation, corresponding to the three gradual turns in the simulation. During these turns, the heavy rain condition performs markedly worse, with errors reaching approximately 10 degrees. In contrast, the steering angles for the derained and clear scenarios remain closely aligned, demonstrating the positive impact of deraining on steering performance. Similarly, Figure \ref{fig:Fig14} highlights that steering performance under light rain conditions is significantly better than under heavy rain, with errors reduced to within 5 degrees. This improvement further emphasizes the detrimental effects of heavy rain on steering accuracy and the potential of deraining to mitigate these challenges.

\begin{figure}
    \centering    \includegraphics[width=0.5\linewidth]{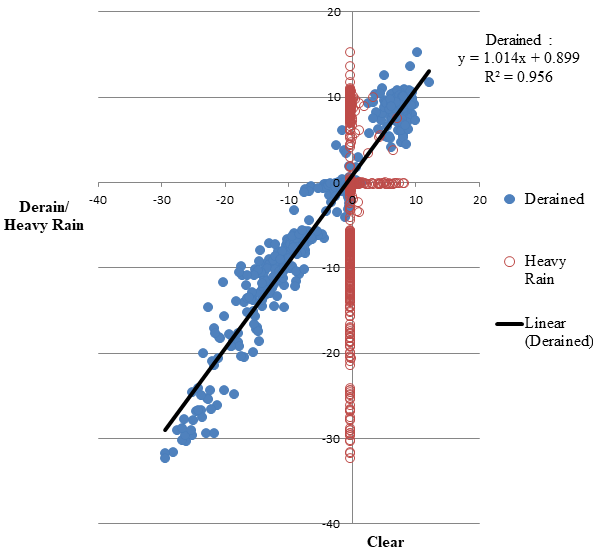}
    \caption{Steering performance: Derained vs heavy rain regression plot.}
    \label{fig:Fig15}
\end{figure}

\begin{figure}
    \centering    \includegraphics[width=0.5\linewidth]{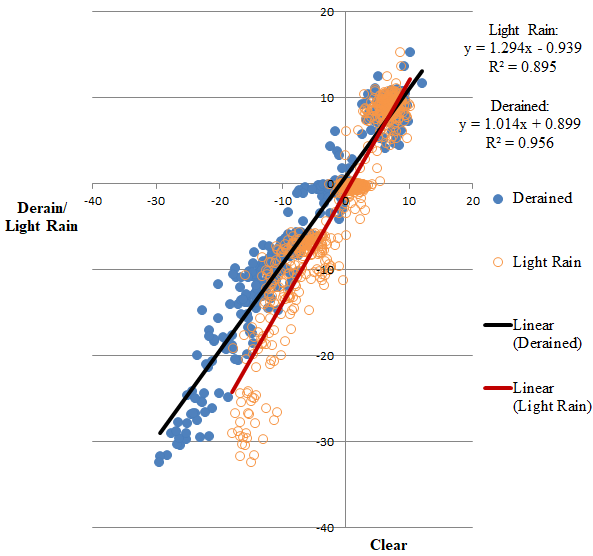}
    \caption{Steering performance: Derained vs light rain regression plot.}
    \label{fig:Fig16}
\end{figure}

To further illustrate the effects of deraining, Figures \ref{fig:Fig15} and \ref{fig:Fig16} present regression plots of predicted steering angles under derained conditions compared to those under corresponding clear conditions, for heavy rain and light rain scenarios, respectively. The vertical axis represents steering angles in clear conditions, while the horizontal axis represents steering angles under derained or rainy conditions. Each plot includes a regression line along with the corresponding \textit{R\textsuperscript{2}} value. As shown in Figure \ref{fig:Fig15}, the regression for Clear vs. Derained conditions achieves an\textit{ R\textsuperscript{2}} value of 0.956, indicating a strong correlation. In contrast, the Clear vs. Heavy Rain regression shows nearly no correlation, as evidenced by the majority of points clustering along the vertical axis. This highlights that in heavy rain, the predicted steering angles fail to respond to curvy road segments. In essence, the vehicle "misses" visual cues in heavy rain, resulting in it continuing straight instead of turning as needed. Figure \ref{fig:Fig16} illustrates a similar comparison for light rain conditions, where the\textit{ R\textsuperscript{2}} value is 0.895, lower than that of the derained conditions. The steeper regression line for light rain indicates a tendency for under-predicted steering angles, meaning the vehicle turns less than necessary in these conditions. These results underscore the effectiveness of our deraining model in improving steering performance, thereby enhancing the safety and reliability of autonomous vehicles in rainy weather.

\section{Conclusions}
Rain presents a formidable challenge for AV navigating roadways as rain streaks can severely impair camera-based objects and feature detection systems employed by AV. Addressing this issue is essential for enhancing AV performance and safety.

We adopted a data-centric approach and introduced two novel batching schemes, STSB and STRB, to improve deraining performance, comparing them to the conventional RTRB batching scheme. STSB paired sequential images both in time and batch, while STRB paired sequential images in time but randomized them across batches. Our results demonstrated that STRB outperformed STSB, primarily due to its ability to incorporate diverse scenes across batches, reducing overfitting and bias compared to sequential scene batching. Additionally, STRB's use of sequential image pairs in time enabled it to better capture dynamic rain features over relatively static road scenes in successive frames. These advantages were evident in reduced training, validation, and testing losses, as well as in superior visual quality of derained images produced by STRB.

The encoder-decoder model developed in this work extended the DCGAN architecture to handle higher-resolution images and incorporates skip-concatenation operations inspired by U-Net. This enabled context-aware image generation, effectively removing rain streaks and achieving weather style transfer. Visual comparisons of rainy, ground-truth (clear), and derained images confirmed the model’s ability to remove rain streaks while transforming rainy scenes into clear, sunny conditions. Compared to other methods, the proposed model demonstrated significantly superior performance. The practical benefits of the deraining model were quantitatively validated using PilotNet to predict AV steering angles on a highway section. Under heavy rain conditions, the AV lost steering control, deviating significantly from the ground-truth steering angles recorded under clear conditions. While light rain improved steering performance, it was under derained conditions that the steering angles closely matched those of clear weather, achieving an \textit{R\textsuperscript{2}} value of 0.956. These results provide robust evidence of the model’s effectiveness in enhancing visibility and improving AV control in rainy conditions.

In conclusion, this study highlights the potential of a data-centric approach combined with deep learning models for joint image deraining and weather style transfer. While the model demonstrates strong performance, certain limitations remain. It addresses the removal of rain steaks but does not account for other weather-related challenges, such as raindrops on the windshield or splashes from preceding vehicles, which may further complicate visibility. Additionally, the simplicity of its architecture limits its ability to preserve finer image details. The use of CARLA-generated datasets, while effective, may not fully capture the diversity of real-world conditions.

Future research could expand the scope to address a wider range of weather conditions, explore architectural enhancements, integrate sequential image frame modeling, and incorporate real-world driving datasets to improve robustness and adaptability. Furthermore, diffusion models \cite{rombach2022high,ho2020denoising} hold potential for real-time applications as their computational efficiency continues to improve, warranting further investigation. It is also important to note that we primarily focus on evaluating the efficacy of deraining in enhancing vehicle steering performance. However, the model holds potential to benefit other critical self-driving tasks, such as object detection, which should be explored in future work.

\bibliographystyle{unsrt}  
\bibliography{references}

\end{document}